\documentclass[conference]{IEEEtran}

\usepackage{balance}  
\usepackage{graphics} 
\usepackage{times}    
\usepackage{url}      
\usepackage{cite}
\usepackage{amsmath,amssymb,amsfonts}
\usepackage{algorithm}
\usepackage{algpseudocode}
\usepackage{mathtools}
\usepackage{bm}
\usepackage[table,xcdraw]{xcolor}

\makeatletter
\def\url@leostyle{%
  \@ifundefined{selectfont}{\def\UrlFont{\sf}}{\def\UrlFont{\small\bf\ttfamily}}}
\makeatother
\def\pprw{8.5in}
\def\pprh{11in}

\begin{document}

\title{Does it matter how well I know what you're thinking?
	 Opponent Modelling in an RTS game}

\author{\IEEEauthorblockN{James Goodman and Simon Lucas}
	\IEEEauthorblockA{Game AI Research Group\\
		School of Electronic Engineering and Computer Science\\
		Queen Mary University of London \\
		Email: \{james.goodman$|$simon.lucas\}@qmul.ac.uk}
	}
\IEEEoverridecommandlockouts
\IEEEpubid{\begin{minipage}{\textwidth}\ \\[12pt]
		000-0-0000-0000-0/00/\$0.00 \copyright 2020 IEEE
\end{minipage}}
\maketitle

\begin{abstract}
Opponent Modelling tries to predict the future actions of opponents, and is required to perform well in multi-player games.
There is a deep literature on learning an opponent model, but much less on how accurate such models must be to be useful.
We investigate the sensitivity of Monte Carlo Tree Search (MCTS) and a Rolling Horizon Evolutionary Algorithm (RHEA) to the accuracy of their modelling of the opponent in a simple Real-Time Strategy game. 
We find that in this domain RHEA is much more sensitive to the accuracy of an opponent model than MCTS.
MCTS generally does better even with an inaccurate model, while this will degrade RHEA's performance.
We show that faced with an unknown opponent and a low computational budget it is better not to use any explicit model with RHEA, and to model the opponent's actions within the tree as part of the MCTS algorithm.
\end{abstract}


\section{Introduction}
When playing a game, or acting in any environment in which other players are present, we would intuitively expect that knowing what the other players are going to do will be helpful in deciding what actions to take. This is true whether their actions are taken in response to our actions, or simply as they interact with the environment regardless of what we decide to do. We need an \emph{opponent model}.

It is also intuitive that if our opponent model is inaccurate, we will do less well. For example if we think our opponent will attack if they have a 3:1 advantage, but in fact they will only attack at 10:1 odds then we will likely play more defensively than optimal. Conversely, if they will actually attack at 2:1 odds then the defences we expect to inhibit an assault will fail to do so.

In a 2-player zero sum game we can theoretically fall-back on the concept of a Nash Equilibrium, which assumes a perfectly rational opponent, although this also reduces our potential to exploit a sub-rational agent. Calculating a Nash Equilibrium strategy may be straightforward in a simple normal-form game, but usually becomes intractable in an extensive form one \cite{Gilpin_Hoda_Pena_Sandholm_2007}. Although there have been major successes here, as with the Counterfactual Regret Minimisation algorithms that approximate a Nash Equilibrium strategy in Texas Hold'em poker, these require very large amounts of pre-processing and domain-specific reductions of the state-space \cite{Johanson_Bard_Burch_Bowling_2012}.

Real Time Strategy (RTS) games have very large branching factors and the potential for simultaneous actions by both players and also by different units of one player. This makes calculation of a Nash Equilibrium for every move infeasible.
Any opponent model also requires some computational time. The more sophisticated our model, the less time we have to make our own decision. On the assumption that in most applications there will only be a limited computational budget available, we also need to trade-off between these. It may be that an inaccurate but cheap opponent model provides better results in actual play than a perfectly accurate but expensive one. 
At the lowest computational limit we have a `DoNothing' opponent model that never does anything.
This computational consideration is important in commercial games, in which we generally cannot leave the human player(s) waiting while the AI calculates its move, and CPU cycles have to be shared with such distractions as graphics, sound and physics engines.

In this paper we investigate the impact of varying the opponent model in a simple RTS-style game, Ground War, developed as a test-bed for ground-based military simulations. 
We investigate two popular Statistical Forward Planning algorithms that make use of a forward model to plan the next action; Monte Carlo Tree Search (MCTS) and Rolling Horizon Evolutionary Algorithms (RHEA).
In each case we ask whether a simple opponent model with minimal computational budget can robustly improve the quality of play, and hence the verisimilitude of simulation results. 
By `robustly' we specifically mean that the opponent model should be helpful against a variety of different actual opponents, and not just those for which the model is perfectly accurate. 
Work with statistical forward planning using a learned model in the Game of Life and Sokoban has shown that the learned environmental model need to be quite accurate to be at all useful \cite{Dockhorn_Lucas_Volz_Bravi_Gaina_Perez-Liebana_2019, Lucas_Dockhorn_Volz_Bamford_Gaina_Bravi_Perez-Liebana_Mostaghim_Kruse_2019}. 
On the other hand recent work emphasised that learned forward models could lead to better outcomes if the model was trained to model what was relevant to gaining reward as opposed to trying to model the full state transitions \cite{schrittwieser2019mastering}.

We wish to determine how opponent model fidelity affects the game playing performance of statistical forward planning algorithms.  We show that MCTS is more robust to using an incorrect opponent model than RHEA.

\section{Background} \label{Background}
\subsection{Opponent Modelling}
Opponent Modelling is where an agent A models what other agents in the environment will do in response to stimuli, where such stimuli include A's own actions.
While in this work it is accurate to refer to `opponent' models, more generally these are `other agent' models, and are just as important in co-operative and semi-cooperative domains.

A few broad categories of opponent model can be distinguished, and see \cite{vandenHerik_Donkers_Spronck_2005, Albrecht_Stone_2018} for more detailed surveys:
\begin{enumerate}
	\item{Game Theoretic. One approach is to assume the opponent is perfectly rational, and seek to find a Nash Equilibrium strategy (or approximation to), for example using counterfactual regret in Poker \cite{Zinkevich_Johanson_Bowling_Piccione_2008, Johanson_Bard_Burch_Bowling_2012}. This guarantees that we cannot be exploited by the other side, but can be very computationally demanding even when tractable.}
	\item{Theory of Mind. Broadly this covers any approach which assumes that the other agents are also modelling us, and that to model their actions successfully it is necessary to also model their model of us (with a theoretically infinite recursion). Examples are the nested cognitive hierarchy of \cite{Stahl_Wilson_1995, Camerer_Ho_Chong_2004}, or the Recursive Modelling Method of \cite{Gmytrasiewicz_Durfee_1995}}
	\item{Own policy. We assume the other agent is using the same algorithm that we are. A good example is MCTS where we model the other player's actions in the tree \cite{Browne_MCTS_2012}. Classic minimax-search through the game-tree is another. In this case we can also use an estimate of the evaluation function that the opponent is using, which may be different to the one we use, especially if the game is asymmetric in any way \cite{Carmel_Markovitch_1996, vandenHerik_Donkers_Spronck_2005}.}
	\item{Heuristic. A hand-crafted `expert' policy that specifies the action to take in any situation has the advantage of reducing computational overhead, but requires domain knowledge and will likely be able to represent a less flexible policy-space compared to the previous categories, especially if the heuristic is kept simple, both in terms of coding investment and time to execute. For example a heuristic policy used as a correctly-specified opponent model is found to significantly improve performance in MCTS~\cite{Walton-Rivers_Williams_Bartle_Perez-Liebana_Lucas_2017}.
	This Heuristic can also be learned off-line from play-traces of human games, and/or be used to augment another approach, for example as a leaf evaluation function in tree search \cite{Mizukami_Tsuruoka_2015, rebstock2019learning}.	}
	\item{Adaptive Model. A policy can be learned/adapted from an opponents observed moves as the game progresses. Approaches include Fictitious Play, or Bayesian updates over distinct heuristics \cite{Albrecht_Stone_2018, Shum_Kleiman-Weiner_Littman_Tenenbaum_2019}.}
	\item{Environmental. In Multi-agent Reinforcement Learning the policy of actions taken by agents is subsumed as part of the environment \cite{Tan_1993}. This is not quite the same as no opponent model, as the opponent does act, but this is seen as a change to the environment and implicitly incorporated into the resultant learned policy.}
\end{enumerate}

As our objective here is to model our opponent with minimal computational overhead we use a Heuristic approach here, plus in the case of MCTS we use an `Own Policy' approach that assumes the opponent is also using MCTS and model their actions in the same tree used for our own actions.

There is a deep literature learning an opponent model, but much less on the impact of getting this model wrong, or even slightly inaccurate \cite{Albrecht_Stone_2018}. In adaptive approaches this should be mitigated as the opponent model learns during play from actual opponent moves. In approaches that use a fixed heuristic or an off-line learned policy, then this risk of poor performance is greater.

Similar work to ours was considered when looking for robust play against unknown opponents in the game of Spades \cite{Sturtevant_Bowling_2006}. That found that a correct opponent model (using three exemplars on trick-play) is best, but can lead to major losses if mis-specified. A soft-max combination of the three exemplars was more robust against an unknown opponent.
Their results also show that an incorrect model can sometimes help if it confuses the opponent by causing you to play contrary to what their correct model of you predicts.

\subsection{Statistical Forward Planning}

Monte Carlo Tree Search (MCTS) \cite{Coulom_2006, Chaslot_De_Jong_Saito_Uiterwijk_2006, Browne_MCTS_2012} has been used successfully in many games and other planning environments. It is an anytime algorithm that uses an available time budget to search the forward game tree from the current state. On each iteration four steps are followed:
\begin{enumerate}
\item{Selection. Select an action to take from the current state. If all actions have been selected at least once then the best one is picked using the Upper Confidence for Trees equation \cite{Kocsis_Szepesvari_2006}:
$J(a)=Q(a) + C\sqrt{\frac{\log(N)}{n(a)}}$
The action $a$ with largest $J(a)$ is selected at state. $N$ is the total number of visits (iterations through) the state; $n(a)$ is the number of those visits that then took action$a$; $Q(a)$ is the mean score for all visits to the state that took action $a$; $C$ is a parameter that controls the trade-off between exploitation, and exploration choosing actions with few visits so far. This step is repeated down the tree of game states until a state is reached with previously untried actions.}
\item{ Expansion. Pick one of the untried actions at random, and expand this, creating a new state in the game tree.}
\item{Simulation. From the expanded state, complete a simulation to obtain a final score. In Ground Wars this does not run to the end of the game, but for 100 time-steps.}
\item{Back-propagation. Back-propagate this final score up the tree. Each state records the mean score of all iterations that take a given action from that state as $Q(a)$ that will affect future Selection steps.
Once the available time budget has been used up the action at the root state with either the highest score or most visits is executed in the actual game environment.}
\end{enumerate}
A transposition-table approach is used in the tree with the node defined by the full visible state \cite{Childs_Brodeur_Kocsis_2008}, rather than an Open Loop approach in which the current state is defined by the action sequence taken to reach it \cite{PerezLiebana_Dieskau_Hunermund_Mostaghim_Lucas_2015}.

Rolling Horizon Evolutionary Algorithms (RHEA) \cite{Perez_Samothrakis_Lucas_Rohlfshagen_2013} evolve a sequence of actions (i.e. a plan) by iterating until the available time budget is exhausted (like MCTS it is anytime):
\begin{enumerate}
\item{Generate a starting plan.}
\item{Randomly mutate the current ‘best’ plan.}
\item{Execute the mutated plan using the forward model and calculate a final score for the end state achieved.}
\item{If this is better than the previous ‘best’ plan, then replace the ‘best’ plan and repeat from step 2.}
\end{enumerate}
Once the time budget is exhausted, the first action in the current best plan is executed in the real environment.
In all cases the score used in Ground War is the material advantage over the other side calculated as 5 points per occupied node, and 1 point per surviving force unit (see Section \ref{GroundWar}).

\section{Ground War} \label{GroundWar}

\begin{figure}[!t] 
	\centering
	\includegraphics[width=3in]{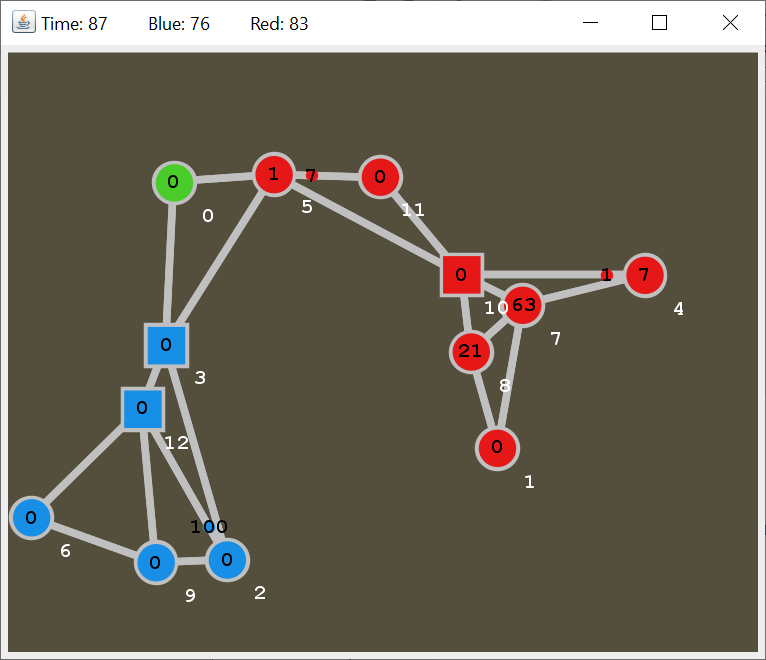}
	\caption{Illustration of a Ground War game in progress between Blue and Red forces. The green node is neutral and not yet occupied by either side. Note the full Blue force of strength 100 is moving from Node 2 to Node 12 (node numbers in white).}
		\label{GWPicture}
\end{figure}
The digital simulation used for this project is ‘Ground War’ (developed on the foundations of the pre-existing `Planet Wars' \cite {Lucas_2018}). This aims to provide a simple, abstract simulation of ground combat with an arc and node map, and with forces able to move from node to node only along connecting arcs as shown in Figure \ref{GWPicture}. Key abstractions are:
\begin{enumerate}
\item{No modelling of specific units. Each force has a size, being a simple numeric value. Each side (Blue and Red) has distinct attributes which apply to all forces of that colour.}
\item{Combat is resolved instantaneously by Lanchester’s Laws, with one force removed in any battle \cite{Engel_1954, Hartley_Helmbold_1995}.}
\item{Once an order has been issued it cannot be cancelled and must be executed in full.}
\end{enumerate}

On top of these abstractions, additional functionality has been built of specific potential interest for wargame simulations to model the impact of fatigue, Command and Control (C2) constraints, Fog of War and Rules of Engagement. Only part of the C2 constraints are relevant to this study, and the remaining functionality is switched off for these experiments.

To represent the cognitive capacity of a human commander and the need to propagate orders to the units concerned a C2 parameter controls a minimum time (in game ticks) that must elapse between orders. 

At any point in time (subject to these C2 constraints) either side may give an order to a force at any node to move to any connected node, and they will then start moving at their speed along the relevant arc. This may be for any percentage of the force currently in the node. For example if Blue has a force of size 45 in a node, then any number up to 45 may be ordered to move. This will split the force, with the remainder staying in the node as a ‘garrison’. 

The Ground War environment is similar in some respects to $\mu$RTS research environment for RTS games \cite{Ontanon_Buro_2015}. It does not have specific unit types, or economic aspects, but has many units that may be given orders simultaneously, and runs in continuous time. These both require adaptations for statistical forward planning from previous work which often has a single agent choosing a single action from a relatively small list at each time step \cite{Perez_Samothrakis_Lucas_Rohlfshagen_2013, Gaina_Lucas_Perez-Liebana_2017, Lucas_Liu_Bravi_Gaina_Woodward_Volz_Perez-Liebana_2019}.

\subsection{Continuous Time}
An order can be issued at any time with constraints imposed by the C2 parameterisation. Discrete time steps, or ‘ticks’, still exist in the simulation, but there is no requirement for an order to be selected for each one. This speeds up the simulation, as the computational overhead of planning is only required on a small subset of ticks.
An order can be one of the following:
\begin{itemize}
\item{LaunchExpedition(X, A, B). Send a force of size X from node A to node B.}
\item{Wait(T). Wait for T ticks. The game forces this if no action may currently be taken due to the C2 constraints.}
\end{itemize}
If the visible situation changes, then a Wait will be interrupted. For example if Red starts moving a force visible to Blue while Blue is Waiting, then Blue will immediately make a new decision in reaction (subject to the C2 constraint). This enables Blue to take a long Wait action until their currently moving forces reach their destinations without losing the ability to react to Red.

\subsection{Action Space}
RHEA evolves a genome that is then translated into a sequence of actions that can be taken in the environment. In contrast MCTS requires a list of actions that are valid in the current state.
For RHEA a genome is a random number in base 10 (for example 35190438313924). This is converted to an action sequence as follows:
\begin{enumerate}
\item{The first digits define A, the node from which an expedition should be launched. If there are fewer than 10 nodes only the first digit is used; if between 10 and 99 then the first two digits are used, and so on.
If A is invalid due to this node not being under the control of the player, or having no garrison, then a Wait order is the default.}
\item{The next digit defines B, the destination. ‘0’ will use the first arc from A, ‘1’ will use the second arc, and so on. (If a map has more than 10 arcs from any node, then two digits will be used here.)}
\item{The next digit defines the proportion of the current garrison force to be sent. This is defined in increments of 10\%, with ‘0’ being 10\% and ‘9’ being 100\%.}
\item{The next digit defines how long to wait after taking the action. This is equal to $d^2$ ticks, where $d$ is the digit value. The C2 constraints apply to set a minimum wait.}
\item{The above steps are then repeated for the next digits in the genome to generate the following action.}
\end{enumerate}

For MCTS we select 20 random actions at each node as the action space when the node is first visited. These are generated from random numeric sequences as above for RHEA until 20 distinct actions are found.
This is a crude form of Action Abstraction \cite{Churchill_Buro_2013, Moraes_Marino_Lelis_Nascimento_2018}, reducing a very large number of possible actions to a small set tractable for forward planning. 

There are more sophisticated methods by which the available actions for MCTS could be obtained, such as progressive unpruning or combinatorial multi-armed bandits (CMAB) \cite{Chaslot_2008, Ontanon_2017}. This approach is used to keep the action space as close as possible to that used for RHEA, and is sufficient to enable the comparison of opponent model behaviour that is our focus.

\section{Experiments} \label{Method}
We investigate two types of opponent model here:
\begin{itemize} 
	\item{Own Policy. In the case of MCTS we can model the opponent's actions using the same algorithm. For each forward simulation in planning we construct a tree for the opponent at the same time, and make their decisions using this. For each simulation we add one node to each tree. For RHEA no `Own Policy' variant is tried; a Rolling Horizon Coevolutionary Algorithm \cite{Liu_Perez-Liebana_Lucas_2016} would be possible, but was found to perform poorly in preliminary experiments.}
	\item{Heuristic. A simple heuristic agent was hand-written which took three parameters:
	\begin{itemize}
		\item{Offence between 1 and 10. The numerical odds required for any Attack to be launched on a node.}
		\item{Defence between 0 and 5. The numerical odds required to not Withdraw when Attacked.}
		\item{Actions. An ordered list of Attack, Withdraw, Reinforce and Redeploy actions (in any order, including omissions). When considering a move, the Heuristic will run through this list and execute the first action that is valid.
			
		\textbf{Attack} is valid if a LaunchExpedition order against an enemy node exists that meets or exceeds the `Offence' parameter odds. 
		
		\textbf{Withdraw} is valid if an enemy attack is inbound on an owned node with fewer defenders than the `Defence' parameter dictates, and a LaunchExpedition will retreat the defenders to an unthreatened node.
		
		\textbf{Reinforce} is valid if the enemy \emph{could} launch an attack on a node that would provoke a Withdraw action. A LaunchExpedition will be executed to send reinforcements from an unthreatened node.
		
		\textbf{Redeploy} is valid if an unthreatened node can send reinforcements to another node that could then launch an Attack action (i.e. an offensive version of `Reinforce').}
	\end{itemize}	
}
\end{itemize}

A set of potentially useful Heuristic agents were obtained using the NTBEA optimisation algorithm \cite{Lucas_Liu_Perez-Liebana_2018} over these three parameters. H0 was handcrafted as a starting point. H1 was then optimised to be able to beat H0. H2 was optimised to play well against RHEA and then H4 against RHEA that used H2 as an opponent model; H3 against MCTS and H5 against MCTS that used H3 as an opponent model. 
The attributes of the resultant Heuristics are listed in Table \ref{Heuristics}.
H1 is a very aggressive player, attacking as long as it has any numerical advantage, while H2-4 will only attack if they have a 10:1 superiority. 
There is little difference between H2 and H3, optimised against RHEA and MCTS respectively; but surprisingly large differences between H4 and H5, optimised against RHEA+H2 and MCTS+H3.

In all cases RHEA and MCTS were given a budget of 50 generations/iterations per decision. 
RHEA used a (1+1) EA with one genome generated per generation with a mutation probability of 0.5, a length of up to 4 actions lasting 100 ticks and a discount factor of 0.999. 
MCTS used a C of 3, a discount factor of 0.999 and a rollout length of up to 100 ticks. 
These parameters were found using NTBEA optimisation with equal time allocated to RHEA and MCTS~\cite{Lucas_Liu_Perez-Liebana_2018}.

With 50 iterations RHEA takes an average of 0.8ms per decision and MCTS twice as long at 1.6ms.
When an opponent model of any type is used by an algorithm these times approximately double. This is because the Ground War forward model does not calculate for each `tick' in turn, but recalculates the game state only in a tick where an action is executed; either at initiation or when a two Forces meet in battle. As a result the computational time required is roughly proportional to the number of player decisions taken and not the number of game ticks that elapse.
The Heuristics take about 800ns to make a decision, so do not impact on the time.
It should be stressed that we are not comparing the performance of RHEA and MCTS as such, but how each algorithm makes use of an (in)accurate opponent model.
The net play level of these algorithms with 50 iterations and 1-2ms per move is subjectively at a `good novice' level appropriate for a simple Game AI; a human player can defeat them once they understand the game.
\begin{table}[]
	\centering
	\begin{tabular}{|l|l|r|r|l|}
		\hline
		Name & Against & Offence & Defence & Actions      \\ \hline
		H0        &                   & 3       & 1.2     & W, A         \\ \hline
		H1        & H0                & 1       & 0.5     & RD, A, W, RF \\ \hline
		H2        & RHEA              & 10      & 1.5     & RD, A, W, RF \\ \hline
		H3        & MCTS              & 10      & 1.0     & RD, W, A, RF \\ \hline
		H4        & RHEA+H2           & 10      & 1.2     & A, W, RF, RD \\ \hline
		H5        & MCTS+H3           & 3       & 0.5     & W, RD, A, RF \\ \hline
	\end{tabular}
\caption{Heuristic Agents used. `Against' is the target against which each was optimised. The Actions are listed in order; \textbf{A}ttack, \textbf{W}ithdraw, \textbf{R}ein\textbf{F}orce and \textbf{R}e\textbf{D}eploy. RHEA+H$?$ means RHEA used with an opponent model of H$?$, where $? \in \{0, 1, 2, 3, 4, 5\}$}
\label{Heuristics}
\end{table}
Two sets of experiments were conducted using these RHEA, MCTS and Heuristic agents. 

\subsection{OM Accuracy experiments}
RHEA and MCTS were tested against a fixed H3 opponent, with an opponent model based on H3 but Offence varied over all integers between 1 and 10, and Defence varied over the range 0.5 to 5.0 at 0.5 intervals. This gives 100 different Heuristic opponents for each match-up, and 2000 games on random maps were run against each.

These experiments were then reversed with a fixed H3 or H0 opponent model, and the actual opponent being varied on the same basis. 
The objective of all these experiments was to see if an opponent model only provides benefit against accurately modelled opponents, and how far this benefit extends.
The H3 and H0 agents from Table \ref{Heuristics} were selected to vary both Offence and the available Actions.

All random maps had 8-10 nodes, and each side started with control of a single random node with a force of 100 units.
\begin{table*}[!t]
	\centering
	\begin{tabular}{|l|l|c|c|c|c|c|c|c|c|c|c|c|c|c|c|}
		\hline
		\multicolumn{2}{|l|}{Agent} & A                                       & B                                       & C                                       & D                                       & E                                       & F                                       & G                                       & H                                       & I                                       & J                                       & K                                       & L                                       & M                                       & Avg                                     \\ \hline
		A         & H0              & 50.0                                  & 44.4                                  & 51.2                                  & 52.0                                  & 51.4                                  & 52.0                                  & 34.6                                  & 47.6                                  & 44.6                                  & 45.0                                  & 45.2                                  & 51.6                                  & 37.2                                  & 46.7                                  \\ \hline
		B         & H1              & 55.6                                  & 50.0                                  & 52.2                                  & 53.0                                  & 51.0                                  & 52.0                                  & 35.2                                  & 32.8                                  & 34.8                                  & 34.6                                  & 34.6                                  & 38.8                                  & 32.6                                  & 42.9                                  \\ \hline
		C         & H2              & 48.8                                  & 47.8                                  & 50.0                                  & 50.2                                  & 48.8                                  & 49.6                                  & 51.8                                  & 47.4                                  & 44.8                                  & 44.2                                  & 42.0                                  & 51.4                                  & 46.4                                  & 47.9                                  \\ \hline
		D         & H3              & 48.0                                  & 47.0                                  & 49.8                                  & 50.0                                  & 48.0                                  & 50.0                                  & 52.0                                  & 46.0                                  & 43.6                                  & 43.0                                  & 43.6                                  & 51.6                                  & 46.8                                  & 47.6                                  \\ \hline
		E         & H4              & 48.6                                  & 49.0                                  & 51.2                                  & 52.0                                  & 50.0                                  & 51.6                                  & 50.4                                  & 47.2                                  & 36.2                                  & 34.6                                  & 35.2                                  & 50.6                                  & 38.8                                  & 45.8                                  \\ \hline
		F         & H5              & 48.0                                  & 48.0                                  & 50.4                                  & 50.0                                  & 48.4                                  & 50.0                                  & 40.4                                  & 46.0                                  & 47.8                                  & 46.8                                  & 47.6                                  & 47.4                                  & 42.6                                  & 47.2                                  \\ \hline
		G         & RHEA+H0         & \cellcolor[HTML]{9AFF99}\textbf{65.4} & 64.8                                  & 48.2                                  & 48.0                                  & 49.6                                  & \cellcolor[HTML]{9AFF99}\textbf{59.6} & 50.0                                  & 49.6                                  & 49.6                                  & 48.0                                  & 49.4                                  & 46.2                                  & 43.6                                  & 51.7                                  \\ \hline
		H         & RHEA+H1         & 52.4                                  & \cellcolor[HTML]{9AFF99}67.2          & 52.6                                  & 54.0                                  & 52.8                                  & 54.0                                  & 50.4                                  & 50.0                                  & 48.4                                  & 47.4                                  & 46.8                                  & 47.4                                  & 41.4                                  & 51.1                                  \\ \hline
		I         & RHEA+H2         & 55.4                                  & \cellcolor[HTML]{9AFF99}65.2          & 55.2                                  & \cellcolor[HTML]{9AFF99}56.4          & \cellcolor[HTML]{9AFF99}63.8          & 52.2                                  & 50.4                                  & 51.6                                  & 50.0                                  & 48.8                                  & 50.4                                  & 47.0                                  & 45.6                                  & 53.2                                  \\ \hline
		J         & RHEA+H3         & 55.0                                  & \cellcolor[HTML]{9AFF99}65.4          & \cellcolor[HTML]{9AFF99}55.8          & \cellcolor[HTML]{9AFF99}\textbf{57.0} & \cellcolor[HTML]{9AFF99}\textbf{65.4} & 53.2                                  & 52.0                                  & 52.6                                  & 51.2                                  & 50.0                                  & 50.6                                  & 48.4                                  & 44.2                                  & 53.9                                  \\ \hline
		K         & RHEA+H4         & 54.8                                  & \cellcolor[HTML]{9AFF99}65.4          & \cellcolor[HTML]{9AFF99}\textbf{58.0} & \cellcolor[HTML]{9AFF99}56.4          & \cellcolor[HTML]{9AFF99}64.8          & 52.4                                  & 50.6                                  & 53.2                                  & 49.6                                  & 49.4                                  & 50.0                                  & 49.0                                  & 45.2                                  & 53.8                                  \\ \hline
		L         & RHEA+RND        & 48.4                                  & 61.2                                  & 48.6                                  & 48.4                                  & 49.4                                  & 52.6                                  & \cellcolor[HTML]{9AFF99}53.8          & 52.6                                  & \cellcolor[HTML]{9AFF99}53.0          & 51.6                                  & 51.0                                  & 50.0                                  & 45.0                                  & 51.2                                  \\ \hline
		M         & RHEA            & \cellcolor[HTML]{9AFF99}62.8          & \cellcolor[HTML]{9AFF99}\textbf{67.4} & 53.6                                  & 53.2                                  & 61.2                                  & \cellcolor[HTML]{9AFF99}57.4          & \cellcolor[HTML]{9AFF99}\textbf{56.4} & \cellcolor[HTML]{9AFF99}\textbf{58.6} & \cellcolor[HTML]{9AFF99}\textbf{54.4} & \cellcolor[HTML]{9AFF99}\textbf{55.8} & \cellcolor[HTML]{9AFF99}\textbf{54.8} & \cellcolor[HTML]{9AFF99}\textbf{55.0} & \cellcolor[HTML]{9AFF99}\textbf{50.0} & \cellcolor[HTML]{9AFF99}\textbf{57.0} \\ \hline
	\end{tabular}
	\caption{Percentage win rates over 500 games on random maps between each pair of RHEA or Heuristic agents. The highest scoring agent against each opponent is in bold, and a green background highlights all agents within a one-tailed 95\% confidence boundary of the best result using an exact Binomial test.}
	\label{RHEAEfficacy}
\end{table*}
\subsection{OM Efficacy experiments}
Two round robin tournaments were run. 
The first includes all six Heuristic agents in Table \ref{Heuristics}, plus RHEA using several of these as opponent models, as well as RHEA without an opponent model (i.e. a "Do Nothing" opponent model) and RHEA with an opponent model that took random actions.
The second includes all Heuristic agents, and otherwise replaces RHEA with MCTS. It also includes an MCTS agent that uses MCTS to model the actions of the opponent; this does not increase the number of calls to the forward model so is `free' if this is the rate-limiting step.
These experiments investigate whether simple opponent models with a low computational overhead can robustly help against more realistic opponents that are outside the modelled policy space.
The same 250 maps are used for all match-ups between agent pairs; each map is played twice with agents alternating sides for a total of 500 games between each pair.

\section{Results} \label{Results}
\subsection{OM Accuracy}\label{RHEAAccuracyResults}

\begin{figure}[!t] 
	\centering
	\includegraphics[width=3in]{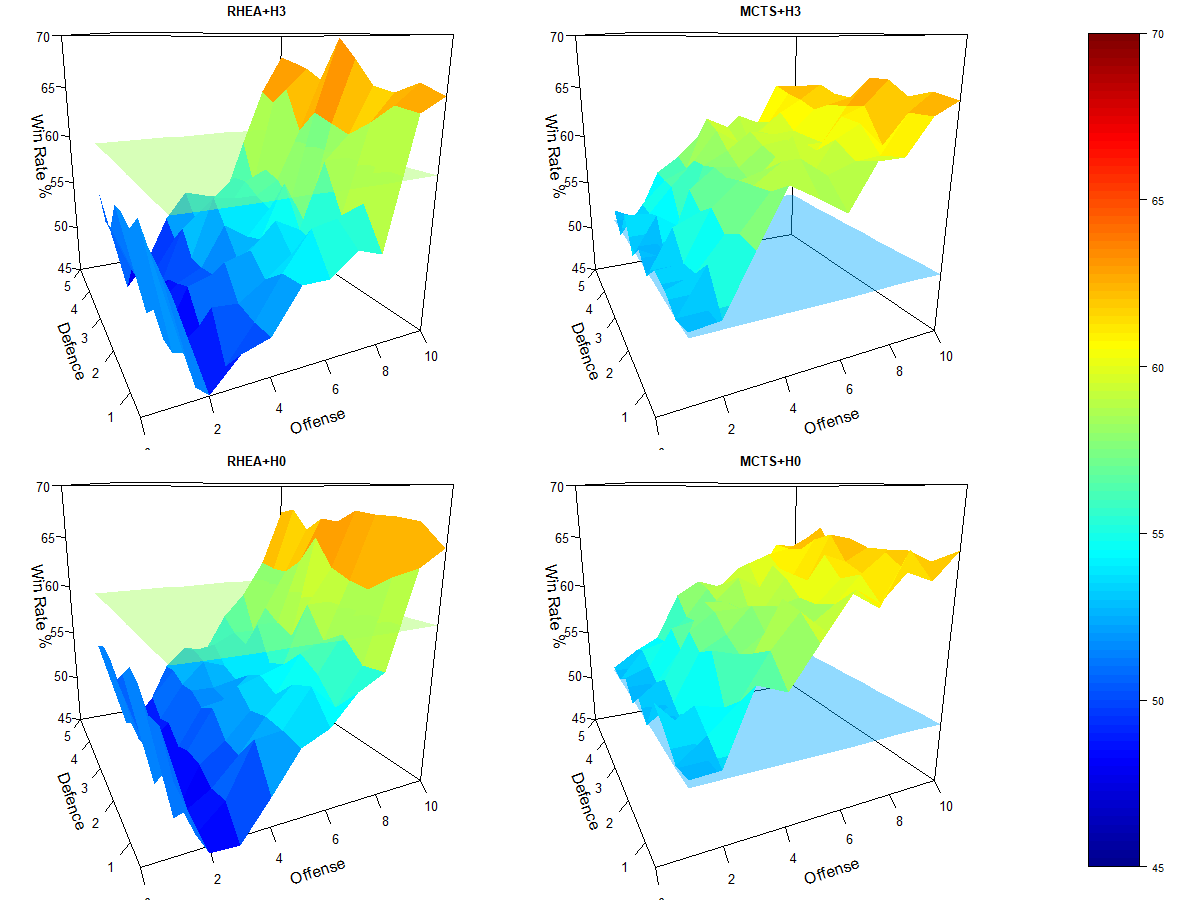}
	\caption{The left-hand side shows the win rate of RHEA+OM against H3 and H0 fixed opponents, where `OM' is based on H3 with Offence between 1 and 10, and Defence between 0.5 and 5.0. The right-hand side shows MCTS+OM on the same basis. 2000 games on random maps were run for each point. The plane in each shows the baseline win rate of RHEA/MCTS with no opponent model against the fixed opponent.}
		\label{RHEAAccuracy}
\end{figure}

\begin{figure}[!t] 
	\centering
	\includegraphics[width=3in]{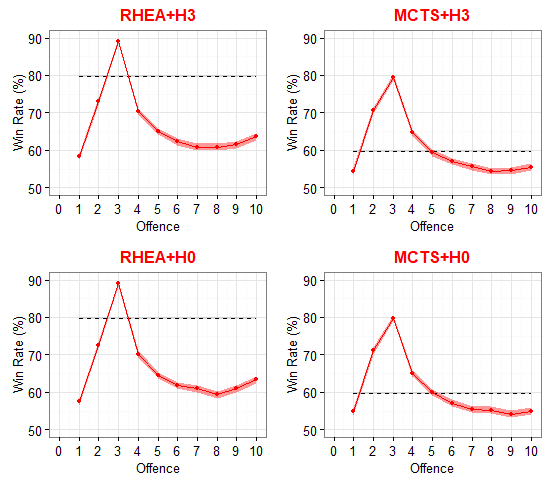}
	\caption{Effect of varying opponent model against a fixed H0 opponent with Offence of 3. The experiments are as in Figure \ref{RHEAAccuracy}, but against H0 as the opponent and the Defence parameter is marginalised out. Shaded regions show 99\% confidence intervals. Solid red line is the win rate of the RHEA/MCTS algorithm and the dotted black line is the baseline performance against H0 of RHEA/MCTS with no opponent model.}
	\label{AccuracyRibbonH0OppModel}
\end{figure}

Figure \ref{RHEAAccuracy} shows that using an opponent model with the correct Offence parameter against H3 leads to better performance than not using an opponent model; there seems to be no significant effect of the Defence parameter. 
This effect falls off rapidly and RHEA with H3 as an opponent model (hereafter abbreviated to RHEA+H3) is only better than RHEA against Heuristic agents with an Offence of either 9 or 10. For any lower values RHEA performs better with no opponent model (i.e. where RHEA assumes that the enemy will sit still and make no new moves at all). 
This is not specific against H3, and the same pattern is shown against H0 (with Offence of 3) in Figure \ref{AccuracyRibbonH0OppModel}, which marginalises out the Defence parameter for clarity.

MCTS+H3 also performs better than MCTS with no opponent model when its actual opponent is close in Offence parameter. 
The fall-off as the opponent model becomes less accurate is less steep than with RHEA, and notably \emph{any} opponent model helps performance.
MCTS with an opponent model is always much better against H3 than vanilla MCTS in striking contrast to the RHEA results; and against H0 the deterioration with a very inaccurate model is small.

\begin{figure}[!t] 
	\centering
	\includegraphics[width=3in]{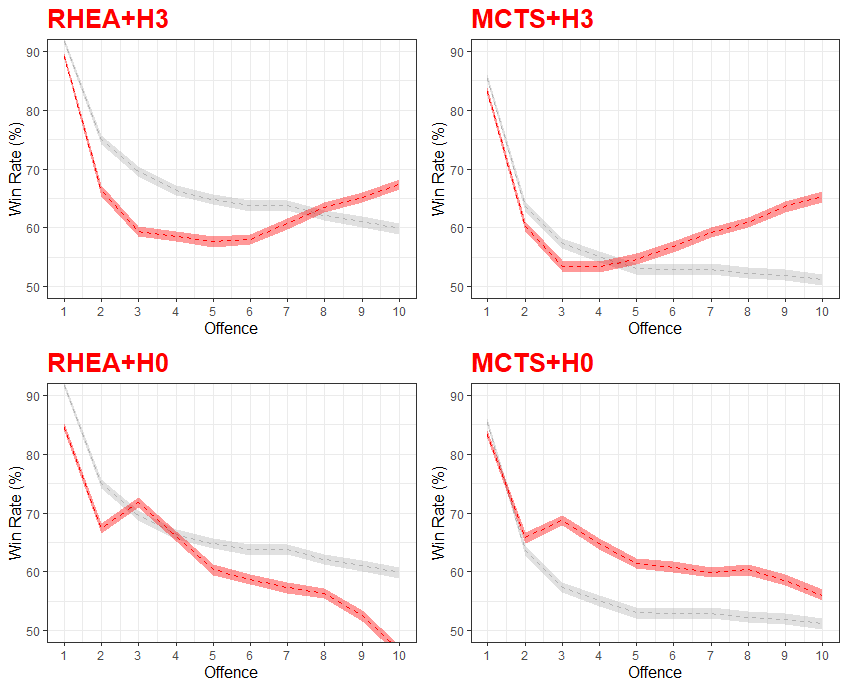}
	\caption{Effect of varying an opponent based on H3 against a fixed opponent model. The top line uses H3 as an opponent model (Offence of 10); the bottom line uses H0 (Offence of 3 and restricted Actions). Shaded regions show 99\% confidence intervals. Solid red line is the win rate of RHEA/MCTS+H3 and the dotted black line is the performance of RHEA/MCTS with no opponent model.}
	\label{AccuracyRibbonH3}
\end{figure}

\begin{figure}[!t] 
	\centering
	\includegraphics[width=3in]{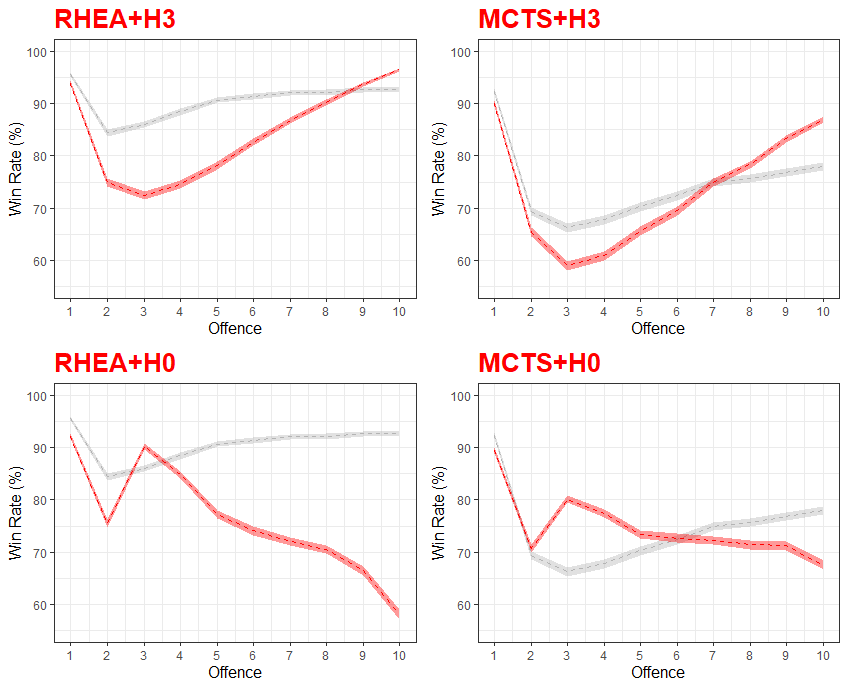}
	\caption{Effect of varying an opponent based on H0 against a fixed opponent model. The experiments and key are as in Figure \ref{AccuracyRibbonH3}}
	\label{AccuracyRibbonH0}
\end{figure}

The effect of keeping the opponent model fixed and varying the actual opponent is shown in Figures \ref{AccuracyRibbonH3} and \ref{AccuracyRibbonH0}.
These show the same pattern. MCTS is almost always better with an opponent model, however inaccurate, while RHEA only benefits from an accurate opponent model. 
The exception to this is when the opponent is based on H0 (Figure \ref{AccuracyRibbonH0}), where RHEA also does better with any opponent model, but not as much as MCTS. H0 never Reinforces or Redeploys, and this reduced action set seems to give the real opponent fewer opportunities to `surprise' the opponent model. 

\subsection{OM Efficacy}
The RHEA results are summarised in Table \ref{RHEAEfficacy}.
RHEA+H0 does best against H0 and H5, the two Heuristic agents with the same Offence rating of 3. RHEA+H3 or RHEA+H4 do well against any Heuristic with an Offence of 10 (H2/H3/H4). 
RHEA with no opponent model only does especially well against H1, as do most RHEA+H? agents. However it does much better overall, and beats every RHEA+H? agent in a head-to-head.
This supports the conclusions from the Accuracy experiments; having a reasonably accurate opponent model improves performance, but hinders performance against opponents for whom the model is not a good fit.

The MCTS results in Table \ref{MCTSEfficacy} show a similar picture. MCTS+H0 does best against H0 and H5 as the most similar Heuristics, and MCTS+H1/3/5 are better against Heuristics than against other MCTS agents.
The results of Section \ref{RHEAAccuracyResults} where MCTS with \emph{any} opponent model does much better against H3 than MCTS without one are replicated in column D of Table \ref{MCTSEfficacy}.
As with RHEA, MCTS with no opponent model generally performs much better against MCTS using an inaccurate opponent model, and MCTS consistently beats MCTS+H?. 
However best overall is MCTS+MCTS, which uses MCTS to model the opponent's actions. This is worse against specific opponents than MCTS using an accurate opponent model, but is best against every single MCTS+H? agent bar one. 
MCTS with no opponent model and MCTS with a random opponent model come a close second overall.
\begin{table*}[]
	\centering
	\begin{tabular}{|l|l|l|l|l|l|l|l|l|l|l|l|l|l|l|l|}
		\hline
		\multicolumn{2}{|l|}{Agent} & \multicolumn{1}{c|}{A}                  & \multicolumn{1}{c|}{B}                  & \multicolumn{1}{c|}{C}                  & \multicolumn{1}{c|}{D}                  & \multicolumn{1}{c|}{E}                  & \multicolumn{1}{c|}{F}                  & \multicolumn{1}{c|}{G}                  & \multicolumn{1}{c|}{H}                  & \multicolumn{1}{c|}{I}                  & \multicolumn{1}{c|}{J}                  & \multicolumn{1}{c|}{K}                  & \multicolumn{1}{c|}{L}                  & M                                       & Avg                                     \\ \hline
		A      & H0            & 50.0                                  & 46.0                                  & 52.0                                  & 51.0                                  & 52.2                                  & 51.2                                  & 36.6                                  & 46.6                                  & 47.8                                  & 46.6                                  & \cellcolor[HTML]{9AFF99}49.0          & 46.6                                  & 46.0                                  & 47.8                                  \\ \hline
		B      & H1            & 54.0                                  & 50.0                                  & 51.2                                  & \cellcolor[HTML]{9AFF99}53.2          & 51.6                                  & 51.6                                  & 36.2                                  & 36.2                                  & 36.4                                  & 35.6                                  & 34.8                                  & 34.6                                  & 35.2                                  & 43.1                                  \\ \hline
		C      & H2            & 48.0                                  & 48.8                                  & 50.0                                  & 50.4                                  & 48.6                                  & 48.6                                  & 48.4                                  & 43.8                                  & 43.0                                  & 43.6                                  & \cellcolor[HTML]{9AFF99}49.8          & \cellcolor[HTML]{9AFF99}49.8          & \cellcolor[HTML]{9AFF99}\textbf{52.0} & 48.1                                  \\ \hline
		D      & H3            & 49.0                                  & 46.8                                  & 49.6                                  & 50.0                                  & 48.4                                  & 49.6                                  & 48.2                                  & 45.8                                  & 43.4                                  & 43.2                                  & 48.0                                  & \cellcolor[HTML]{9AFF99}50.4          & \cellcolor[HTML]{9AFF99}51.0          & 48.0                                  \\ \hline
		E      & H4            & 47.8                                  & 48.4                                  & 51.4                                  & 51.6                                  & 50.0                                  & 51.0                                  & 49.6                                  & 39.8                                  & 41.6                                  & 40.6                                  & 46.2                                  & 45.2                                  & \cellcolor[HTML]{9AFF99}49.8          & 47.2                                  \\ \hline
		F      & H5            & 48.8                                  & 48.4                                  & 51.4                                  & 50.4                                  & 49.0                                  & 50.0                                  & 42.6                                  & 47.4                                  & 47.8                                  & 49.0                                  & 46.6                                  & 45.4                                  & 48.2                                  & 48.1                                  \\ \hline
		G      & MCTS+H0       & \cellcolor[HTML]{9AFF99}\textbf{63.4} & \cellcolor[HTML]{9AFF99}63.8          & 51.6                                  & 51.8                                  & 50.4                                  & \cellcolor[HTML]{9AFF99}\textbf{57.4} & 50.0                                  & 50.0                                  & 51.2                                  & 49.8                                  & 45.8                                  & 45.8                                  & 40.4                                  & 51.6                                  \\ \hline
		H      & MCTS+H1       & 53.4                                  & \cellcolor[HTML]{9AFF99}63.8          & \cellcolor[HTML]{9AFF99}56.2          & \cellcolor[HTML]{9AFF99}54.2          & \cellcolor[HTML]{9AFF99}\textbf{60.2} & 52.6                                  & 50.0                                  & 50.0                                  & 50.4                                  & 50.0                                  & 44.4                                  & 45.0                                  & 41.8                                  & 51.7                                  \\ \hline
		I      & MCTS+H3       & 52.2                                  & \cellcolor[HTML]{9AFF99}63.6          & \cellcolor[HTML]{9AFF99}\textbf{57.0} & \cellcolor[HTML]{9AFF99}56.6          & \cellcolor[HTML]{9AFF99}58.4          & 52.2                                  & 48.8                                  & 49.6                                  & 50.0                                  & 48.6                                  & 43.6                                  & 45.8                                  & 40.8                                  & 51.3                                  \\ \hline
		J      & MCTS+H5       & 53.4                                  & \cellcolor[HTML]{9AFF99}64.4          & \cellcolor[HTML]{9AFF99}56.4          & \cellcolor[HTML]{9AFF99}\textbf{56.8} & \cellcolor[HTML]{9AFF99}59.4          & 51.0                                  & 50.2                                  & 50.0                                  & 51.4                                  & 50.0                                  & 44.6                                  & 46.0                                  & 40.6                                  & 51.9                                  \\ \hline
		K      & MCTS+RND      & 51.0                                  & \cellcolor[HTML]{9AFF99}65.2          & 50.2                                  & 52.0                                  & 53.8                                  & 53.4                                  & 54.2                                  & \cellcolor[HTML]{9AFF99}55.6          & \cellcolor[HTML]{9AFF99}56.4          & 55.4                                  & \cellcolor[HTML]{9AFF99}50.0          & \cellcolor[HTML]{9AFF99}49.2          & 47.6                                  & 53.4                                  \\ \hline
		L      & MCTS          & 53.4                                  & \cellcolor[HTML]{9AFF99}\textbf{65.4} & 50.2                                  & 49.6                                  & 54.8                                  & \cellcolor[HTML]{9AFF99}54.6          & 54.2                                  & \cellcolor[HTML]{9AFF99}55.0          & 54.2                                  & 54.0                                  & \cellcolor[HTML]{9AFF99}50.8          & \cellcolor[HTML]{9AFF99}50.0          & 47.8                                  & 53.4                                  \\ \hline
		M      & MCTS+MCTS     & 54.0                                  & \cellcolor[HTML]{9AFF99}64.8          & 48.0                                  & 49.0                                  & 50.2                                  & 51.8                                  & \cellcolor[HTML]{9AFF99}\textbf{59.6} & \cellcolor[HTML]{9AFF99}\textbf{58.2} & \cellcolor[HTML]{9AFF99}\textbf{59.2} & \cellcolor[HTML]{9AFF99}\textbf{59.4} & \cellcolor[HTML]{9AFF99}\textbf{52.4} & \cellcolor[HTML]{9AFF99}\textbf{52.2} & \cellcolor[HTML]{9AFF99}50.0          & \cellcolor[HTML]{9AFF99}\textbf{54.5} \\ \hline
	\end{tabular}
\caption{Percentage win rates over 500 games on random maps between each pair of MCTS or Heuristic agents. The highest scoring agent against each opponent is in bold, and a green background highlights all agents within a one-tailed 95\% confidence boundary of the best result using an exact Binomial test.}
\label{MCTSEfficacy}
\end{table*}

\subsection{RHEA Horizon Length}
\begin{figure}[!t] 
	\centering
	\includegraphics[width=2in]{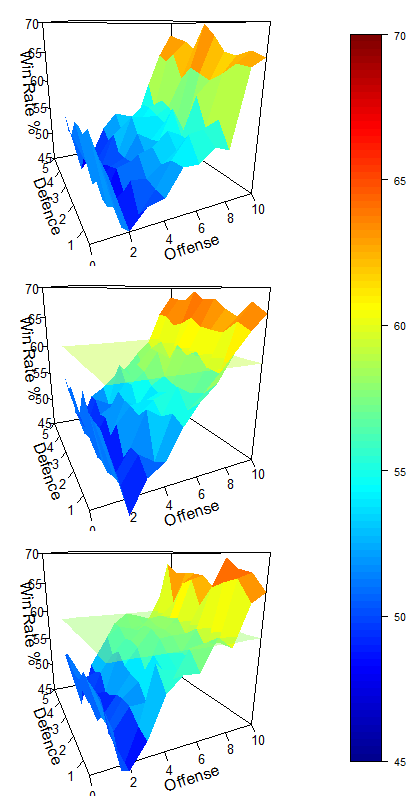}
	\caption{Impact of changing RHEA sequence length (the number of forward actions in each plan). The top-most graph is for a sequence length for 4 actions as in Figure \ref{RHEAAccuracy}. The middle graph is for a sequence length of 2 actions, and the bottom-most for just a single action.}
		\label{RHEASequenceLength}
\end{figure}
Ground War is a 2-player zero sum game, and what the other player does very emphatically affects our score. In this environment we expect modelling the opponent action to be helpful, and the fact that RHEA is often more successful if it assumes the opponent does nothing is counter-intuitive and in need of explanation.
The MCTS result is more congruent with our a priori expectation.

The results suggest that RHEA is particularly sensitive to a poor opponent model in comparison to MCTS.
One hypothesis is that RHEA is more sensitive to a poor assumption because it always plans forward for $x$ actions (where $x=4$ in Figure \ref{RHEAAccuracy}), and applies equal amounts of computation to each action.
MCTS focuses more of its budget on the first action taken as it builds up the game tree over iterations; every single iteration will make a choice in the tree for the first action, but only the last few iterations of the 50, if any, will make a decision for the third action.
This means that RHEA may generate a plan that is dependent on anticipated moves by the opponent over longer period of time relative to MCTS. 
In this case searching less far forward in time can be beneficial.

To test this hypothesis we ran repeat experiments with RHEA that have a reduced horizon of 1 or 2 actions. These results are shown in Figure \ref{RHEASequenceLength}.
Even when RHEA is only planning one action forward there is still a steep fall-off in performance for an incorrect opponent model to below a baseline with none. There is however a plateau for an opponent model Offence rating between 4 and 8 that is not significantly different to the baseline.
This lends some support to our hypothesis but it is clearly far from sufficient to explain the full difference between the effect on an opponent model in RHEA and MCTS.
A full understanding of this is a key area for future work.

\section{Conclusion and Future Work} \label{Conclusions}
We have used a parameterised Heuristic opponent in a simplified RTS, Ground War, to investigate the effect of accuracy of an opponent model for performance of MCTS and RHEA agents with a small computational budget.
For both algorithms we have varied the actual opponent while keeping the opponent model constant, and vice versa. We repeated this for two different constant opponents and opponent models (H0 and H3) picked to be very different from each other.

We have shown that in this domain having an accurate opponent model in statistical forward planning is beneficial and improves performance. With RHEA this benefit rapidly falls away as the opponent model becomes less accurate and the experimental results suggest that using no opponent model at all (assuming that the opponent never acts) can be the best approach if we are uncertain about the their actual policy.

MCTS also benefits from an accurate opponent model, and here the fall-off is much shallower. An opponent model is usually still beneficial even if quite inaccurate. However, modelling the opponent within the MCTS tree itself is much more robust and is preferable if the opponent policy is unknown.

In this work we do not adapt the opponent model based on observation of actions taken during the game so far. This is a common approach 
and to the extent that it improves the accuracy of the opponent model should be beneficial.
The research question of how accurate an opponent model needs to be is just as valid in this adaptive method given that any learned model is still constrained to a policy-space that may not include the actual opponent.

Further work is especially needed to understand the precise origin of the difference in behaviour of RHEA and MCTS with an opponent model, and investigate this effect in other games beyond the one used here.



\section{Acknowledgments}
This work was funded by the EPSRC CDT in Intelligent Games and Game Intelligence (IGGI) EP/S022325/1.

%
%
%
%
%
\balance

\bibliographystyle{acm-sigchi}
\bibliography{bibliography}
\end{document}